\pdfoutput=1

\documentclass[11pt]{article}

\usepackage[preprint]{acl}

\usepackage{times}
\usepackage{latexsym}
\usepackage{tabularx}
\usepackage{longtable}
\usepackage{booktabs}
\usepackage[T1]{fontenc}

\usepackage[utf8]{inputenc}

\usepackage{microtype}

\usepackage{inconsolata}

\usepackage{graphicx}
\usepackage{enumitem}
\usepackage{amsmath}
\usepackage{amssymb} 
\usepackage{booktabs}
\usepackage{tcolorbox}
\usepackage{colortbl}
\usepackage{color}
\usepackage{xcolor}
\usepackage{mdframed}
\usepackage{supertabular}

\definecolor{NavyBlue}{rgb}{0.0, 0.0, 0.5}
\definecolor{BlueGreen}{rgb}{0.0, 0.5, 0.5}
\definecolor{Plum}{rgb}{0.5, 0.0, 0.5}
\definecolor{ForestGreen}{rgb}{0.0, 0.26, 0.15}
\definecolor{Red}{rgb}{1.0, 0.0, 0.0}

\usepackage{listings}
\usepackage{xcolor}
\lstdefinelanguage{json}{
    basicstyle=\ttfamily,
    numbers=none,
    numberstyle=\tiny,
    stepnumber=1,
    breaklines=true,
    showstringspaces=false,
    frame=none,
    backgroundcolor=\color{white},
    literate=
     *{0}{{{\color{black}0}}}{1}
      {1}{{{\color{black}1}}}{1}
      {2}{{{\color{black}2}}}{1}
      {3}{{{\color{black}3}}}{1}
      {4}{{{\color{black}4}}}{1}
      {5}{{{\color{black}5}}}{1}
      {6}{{{\color{black}6}}}{1}
      {7}{{{\color{black}7}}}{1}
      {8}{{{\color{black}8}}}{1}
      {9}{{{\color{black}9}}}{1}
      {:}{{{\color{black}:}}}{1}
      {,}{{{\color{black},}}}{1}
      {"}{{{\color{black}"}}}{1},
}

\title{LLMSR@XLLM25: Less is More: Enhancing Structured Multi-Agent Reasoning via Quality-Guided Distillation}

\author{
 \textbf{Jiahao Yuan\textsuperscript{1}},
 \textbf{Xingzhe Sun\textsuperscript{1}},
 \textbf{Xing Yu\textsuperscript{1}},
\\
  \textbf{Jingwen Wang\textsuperscript{1}},
  \textbf{Dehui Du \textsuperscript{1}}\thanks{Corresponding author.},
 \textbf{Zhiqing Cui\textsuperscript{2}},
  \textbf{Zixiang Di\textsuperscript{1}}
\\
\\
 \textsuperscript{1}East China Normal University,  \quad
 \textsuperscript{2}University of Reading
\\
 \href{mailto:51275900024@stu.ecnu.edu.cn}{\texttt{51275900024@stu.ecnu.edu.cn}}, \quad
  \href{mailto:dhdu@sei.ecnu.edu.cn}{\texttt{dhdu@sei.ecnu.edu.cn}}
}

\begin{document}
\maketitle
\begin{abstract}
The LLMSR@XLLM25 formulates a low-resource structural reasoning task that challenges LLMs to generate interpretable, step-by-step rationales with minimal labeled data. We present \textbf{Less is More}, the third-place winning approach in the LLMSR@XLLM25, which focuses on structured reasoning from only 24 labeled examples. Our approach leverages a multi-agent framework with reverse-prompt induction, retrieval-augmented reasoning synthesis via GPT-4o, and dual-stage reward-guided filtering to distill high-quality supervision across three subtasks: question parsing, CoT parsing, and step-level verification. All modules are fine-tuned from \texttt{Meta-Llama-3-8B-Instruct} under a unified LoRA+ setup. By combining structure validation with reward filtering across few-shot and zero-shot prompts, our pipeline consistently improves structure reasoning quality. These results underscore the value of controllable data distillation in enhancing structured inference under low-resource constraints. Our code is available at \url{https://github.com/JhCircle/Less-is-More}.
\end{abstract}

\section{Introduction}

Structured reasoning tasks—such as decomposing a question into logical constraints or verifying a chain of deductions—pose unique challenges for large language models (LLMs) \cite{zhang2025sr}, especially under extreme low-resource conditions. The LLMSR@XLLM25 targets this very challenge, requiring participants to generate interpretable and verifiable reasoning processes from only \textbf{24} labeled examples. Each instance involves four intertwined subtasks: extracting question conditions (\textit{Question Parsing}), identifying reasoning steps and their justifications (\textit{CoT Parsing}), and validating whether evidence supports each inferred statement (\textit{CoT Statement and Verification}).

This setting presents two core challenges: (1) insufficient labeled data to fine-tune high-capacity models, and (2) the need to maintain step-level consistency and logical coherence across multiple reasoning modules. Prior work on CoT-style prompting typically relies on large-scale instruction tuning or heuristic prompting, which falters when supervision is scarce and structural granularity is essential.

To tackle these challenges, we introduce \textbf{Less is More}—a structured multi-agent framework that transforms minimal supervision into high-quality training signals through three key stages: (i) \textbf{prompt induction} via reverse thinking~\cite{yuan2024reversal,zhou2022large} to derive task-specific instructions; (ii) \textbf{retrieval-augmented reasoning synthesis} with GPT-4o to generate contextually grounded annotations at scale \cite{ram2023context,zhao2024knn}; and (iii) \textbf{dual-stage filtering}, which integrates lightweight structural pruning and reward-based selection to ensure semantic fidelity. Each reasoning module is fine-tuned independently from \texttt{Meta-Llama-3-8B-Instruct} on distilled CoT data generated by GPT-4o~\cite{wei2022chain,zhou2023thread}, enabling modular, interpretable reasoning under low-resource settings.

Our approach ranked third in such shared task, outperforming several strong baselines. Through detailed experiments across diffrent reward-filtering stratgies, we show that data \textit{quality}—not quantity—is the key to enhancing structured reasoning. This highlights the promise of controllable, quality-centric distillation in advancing LLM reasoning under real-world data scarcity.

\section{Methodology}

We present \textbf{Less is More}, a structured multi-agent reasoning framework designed to address data scarcity through \textit{quality-guided distillation}. Given only 24 labeled examples in the LLMSR@XLLM25, we construct a scalable pipeline for data synthesis and filtering. 

Our approach focuses on two reasoning subtasks, each treated as an instruction-following generation problem:
\begin{itemize}
\item \textbf{Question Parsing (QP)}: Infers a structured list of reasoning components (e.g., constraints, relations, entities) directly from the natural language question.
\item \textbf{Unified CoT Reasoning (UCoT)}: Constructs a structured reasoning trajectory by first parsing the chain-of-thought into atomic logical statements (CP), followed by stepwise grounding and validation through evidence retrieval (CS) and verification (CV).
\end{itemize}
\begin{figure}[htbp]
    \centering
    \includegraphics[width=\linewidth]{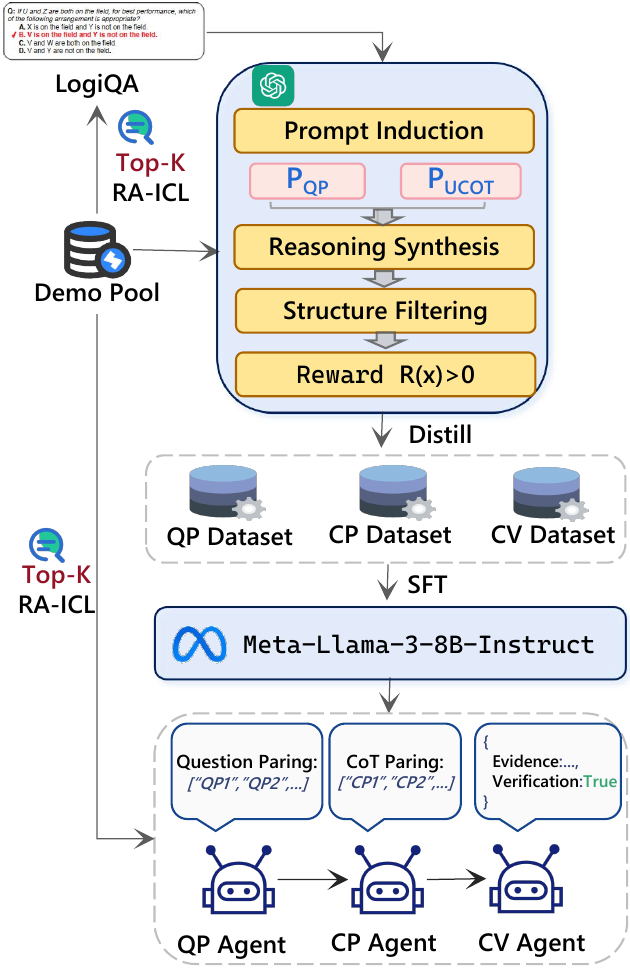}
    \caption{Overview of the \textbf{Less is More} reasoning framework. Training includes reverse-prompt induction, GPT-4o-based synthesis, and reward filtering. Inference deploys fine-tuned agents for question parsing and structured CoT generation.}
    \label{fig:less-is-more}
\end{figure}
Each instance is generated using only a small seed set and guided by reverse prompt induction~\cite{yuan2024reversal}. The full pipeline includes prompt design, reasoning synthesis via retrieval-augmented in-context learning, and reward-based filtering to ensure output quality.

\subsection{Prompt Induction via Reverse Thinking}

To enable instruction-following reasoning under low-resource supervision, we adopt a meta-cognitive prompting strategy following RoT~\cite{yuan2024reversal}. For each subtask $t \in \{QP, UCoT\}$, our goal is to induce task-specific prompts $\mathcal{P}_t$ from a small set of labeled examples, as detailed in Appendix~\ref{app:prompts}.

Let $\mathcal{D}_{\text{seed}}^t = \{(x_i, y_i)\}_{i=1}^{N}$ denote the labeled data for subtask $t$, where $x_i$ is the input and $y_i$ the structured output. We prompt the language model $\mathcal{LLM}$\footnote{We use \texttt{gpt-4o-2024-08-06} with a temperature of 0.1.} with a reverse-thinking instruction $\mathcal{P}_{\text{reverse}}$ and demonstrations from $\mathcal{D}_{\text{seed}}$ to generate an optimal task-specific prompt $\pi_t^*$:

\begin{align}
\Pi 
  &= \mathcal{LLM}(\mathcal{P}_{reverse},\, \mathcal{D}_{seed}) \\
\pi_t^* 
  &= \arg\max_{\pi \in \Pi} \left[ \texttt{S}_{gen}(\pi) + \texttt{S}_{pref}(\pi) \right] \\
\texttt{S}_{gen}(\pi) 
  &= \mathbb{E}_{(x, y) \sim \mathcal{D}_{seed}^t} \left[ \log p_\pi(y \mid x) \right] \\
\texttt{S}_{pref}(\pi) 
  &= \sum_{\pi' \in \Pi \setminus \{\pi\}} \mathbb{I}\left[\pi \succ \pi'\right]
\end{align}

\subsection{Reasoning Synthesis via Retrieval-Augmented ICL}

We use the induced prompts $\{\mathcal{P}_{QP}, \mathcal{P}_{UCoT}\}$ to synthesize structured annotations for unlabeled LogiQA~\cite{liu2021logiqa} instances through a \textbf{retrieval-augmented in-context learning (RA-ICL)} framework inspired by \cite{ram2023context}.

Given a question $x$, we first embed it as $\mathbf{h}_x = f_{\text{enc}}(x)$ using a pretrained encoder.\footnote{\url{https://huggingface.co/BAAI/bge-m3}} Then, we retrieve $k$ similar examples from $\mathcal{D}_{\text{seed}}$ based on cosine similarity:

\begin{equation}
\mathcal{R}(x) = \texttt{TopK}_{x'} \left( \cos(\mathbf{h}_x, \mathbf{h}_{x'}) \mid x' \in \mathcal{D}_{\text{seed}} \right)
\end{equation}

We construct two prompts for each $x$: $\mathcal{P}_{QP}(x)$ for question parsing, and $\mathcal{P}_{UCoT}(x)$ for generating the full reasoning chain with verification. The model then generates:

\begin{align}
\hat{y}_{QP} &= \mathcal{LLM}(\mathcal{P}_{QP}, \mathcal{P}_{QP}(x)) \\
\hat{y}_{UCoT} &= \mathcal{LLM}(\mathcal{P}_{UCoT}, \mathcal{P}_{UCoT}(x))
\end{align}

The output $\hat{y}_{UCoT}$ is a structured json object with fields \texttt{cot\_steps}, each containing a reasoning statement, its textual evidence, and a boolean verification label.

\subsection{Data Filtering via Reward-Based Filtering}
\label{subsec:reward}
To construct a high-quality training set from synthesized CoT annotations, we apply a two-stage filtering process: structure-based pruning followed  by reward-based selection using a fine-tuned reward model inspired by \cite{zhou2023lima,zhou2024less,li2024turning,deng2025less}. This ensures that only structurally valid and semantically meaningful traces are retained fordownstream training.
\setlist{itemsep=0pt, topsep=0pt}
\paragraph{Structural Filtering}: Remove ill-formed or trivial outputs (e.g., malformed JSON, less than two reasoning steps, parsing failures).
\paragraph{Reward-Based Filtering}: we perform \textbf{reward-based filtering} using a top-ranked LLaMA3-based reward model~\footnote{\url{https://huggingface.co/Ray2333/GRM-Llama3.2-3B-rewardmodel-ft}} trained on Reward-Bench~\footnote{\url{https://huggingface.co/spaces/allenai/reward-bench}}. Inspired by prior work~\cite{zhou2023lima,zhou2024less,li2024superfiltering,deng2025less}, we use this model to assess the quality of each reasoning trace under two distinct prompting configurations:

\begin{itemize}
    \item \textbf{Few-shot prompt}: Includes $k$ semantically similar examples retrieved from a demonstration pool along with the synthesized reasoning.
    \item \textbf{Zero-shot prompt}: Uses only the instruction template and the generated reasoning, without any demonstrations.
\end{itemize}
Both prompt variants are formatted as chat-style input-response pairs and passed to the reward model $f_{reward}$. The final reward is defined as the mean of the two scores:
\begin{align}
s_{\text{few}} &= f_{\text{reward}}(\mathcal{P}_{\text{few}}(x),\, \hat{y}) \\
s_{\text{zero}} &= f_{\text{reward}}(\mathcal{P}_{\text{zero}}(x),\, \hat{y}) \\
s_{\text{avg}} &= \frac{1}{2} \left(s_{\text{few}} + s_{\text{zero}}\right) \\
\mathcal{S}(x) &=
\begin{cases}
s_{\text{few}} & \text{(few-shot filtering)} \\
s_{\text{zero}} & \text{(zero-shot filtering)} \\
s_{\text{avg}} & \text{(average-based filtering)}
\end{cases}
\end{align}

We filter examples by thresholding the reward score $\mathcal{S}(x) > 0$, resulting in three filtered subsets based on: (i) few-shot reward, (ii) zero-shot reward, and (iii) their average. Each filtered dataset is stored independently and used to fine-tune the model under the corresponding configuration \footnote{All fine-tuning experiments are conducted on \texttt{meta-llama/Meta-Llama-3-8B-Instruct}, as required by the shared task.}. This enables targeted ablation studies and comparative evaluation of how different reward signals influence downstream performance.

This dual-prompt scoring strategy enables more robust reward estimation by capturing both contextual coherence (few-shot) and general quality (zero-shot), effectively reducing noisy traces and improving training reliability.

\section{Inference Pipeline: Multi-Agent Structured Reasoning}
We deploy a structured inference pipeline that mirrors our modular training architecture, comprising three dedicated agents: \textbf{\textit{Parser}}, \textbf{\textit{Decomposer}}, and \textbf{\textit{Verifier}}, each instantiated as a fine-tuned \texttt{LLaMA3-8B-Instruct} model. These agents are respectively responsible for \textit{question parsing} (QP), \textit{chain-of-thought (CoT) parsing} (CP), and \textit{step-level statement and verification} (CV).

At inference time, given a test instance $x$, we encode it into a dense embedding $\mathbf{h}_x$ using a multilingual encoder\footnote{\url{https://huggingface.co/BAAI/bge-m3}} and retrieve semantically similar exemplars $\mathcal{R}(x)$ from the distillation pool. These demonstrations are reused across all agents via task-specific prompting templates, ensuring consistency and contextual alignment. The full reasoning pipeline unfolds as a cascade of agent interactions:

\begin{align}
\hat{y}_{QP} &= \textsc{Parser}\left(\mathcal{P}_{QP}(x;\, \mathcal{R}(x))\right) \\
\hat{y}_{CP} &= \textsc{Decomposer}\left(\mathcal{P}_{CP}(x,\, CoT;\, \mathcal{R}(x))\right) \\
\hat{e}_{CV} &= \textsc{Verifier}\left(\mathcal{P}_{CV}^{evidence}(x,\, \hat{y}_{CP};\, \mathcal{R}(x))\right) \\
\hat{v}_{CV} &= \textsc{Verifier}\left(\mathcal{P}_{CV}^{verify}(x,\, \hat{y}_{CP},\, \hat{e}_{CV};\, \mathcal{R}(x))\right)
\end{align}

where $\mathcal{P}_{QP}, \mathcal{P}_{CP}, \mathcal{P}_{CV}^{evidence}, \mathcal{P}_{CV}^{verify}$ are prompt construction modules tailored for the Parser, Decomposer, and Verifier respectively (detailed in Appendix~\ref{app:prompts}. \( \hat{y}_{QP} \) denotes the question parsing answer, \( \hat{y}_{CP} \) is the generated answer after cot decomposition, \( \hat{e}_{CV} \) refers to CoT evidence supporting the answer \( \hat{y}_{CP} \), and \( \hat{v}_{CV} \) is the final verification result indicating whether the statement is supported given the evidence.

\section{Experiment}
\subsection{Datasets}
We evaluate the impact of different reward filtering strategies on model performance using the public testsets \footnote{\url{https://huggingface.co/datasets/shuyi-zsy/LLMSR/tree/main/llmsr}} of the LLMSR@XLLM25. Each strategy yields a distinct training dataset, filtered by the corresponding reward signal—\textit{few-shot}, \textit{zero-shot}, or \textit{average-based}—as described in Section~\ref{subsec:reward}. Table~\ref{tab:reward-stats} summarizes the number of training instances retained under each strategy. Notably, the numbers for QP (Question Parsing) and CP (CoT Parsing) are counted at the question and CoT level, where each instance corresponds to a complete reasoning trace. In contrast, the CV (CoT Verification) subtask is formulated as step-level verification, where each reasoning trace is decomposed into multiple verifiable steps. An illustrative example from the synthesized dataset is presented in Appendix~\ref{app:example} to demonstrate the structure and annotation format.

\begin{table}[htbp]
\centering
\small
\setlength{\tabcolsep}{6pt}
\renewcommand{\arraystretch}{1.2}
\begin{tabular}{lrrrr}
\toprule
\textbf{Strategy} & \textbf{Total} & \textbf{QP} & \textbf{CP} & \textbf{CV} \\
\midrule
Original LogiQA       & 7,376 & --    & --    & --    \\
Structure Filtered    & 1,940 & 1,940 & 1,940 & 13,818 \\
0-shot Reward         & 1,309 & 1,309 & 1,309 & 9,434 \\
5-shot Reward         & 1,377 & 1,377 & 1,377 & 9,858 \\
Avg. Reward           & 1,346 & 1,346 & 1,346 & 9,688 \\
\bottomrule
\end{tabular}
\caption{Training set sizes under different filtering strategies. }
\label{tab:reward-stats}
\end{table}

We fine-tune three task-specific models for QP (Question Parsing), CP (CoT Parsing), and CV (CoT Verification) on their respective filtered datasets from our distillation pipeline. Each model is trained independently using \texttt{meta-llama/Meta-Llama-3-8B-Instruct} with LoRA+~\cite{hayou2024lora+} (rank 16, $\alpha = 32$, \texttt{lorap\_lr\_ratio} = 16) via \texttt{ms-swift}\footnote{\url{https://github.com/modelscope/ms-swift}}. Training is conducted for 5 epochs with a learning rate of $2 \times 10^{-5}$, batch size 4 per device, gradient accumulation over 4 steps, and a warmup ratio of 0.03, using two \texttt{NVIDIA A100-80G} GPUs.

\subsection{Results}
\begin{table}[htbp]
\centering
\small
\setlength{\tabcolsep}{1pt}
\renewcommand{\arraystretch}{1.2}
\begin{tabular}{lcccc}
\toprule
\textbf{Setting} & \textbf{Ques. F1} & \textbf{Stmt. F1} & \textbf{Evid. F1} & \textbf{Reason. F1} \\
\midrule
Structure Filtered & 56.87 & 36.72 & 10.80 & 5.20 \\
0-shot Reward      & 62.76 & 38.05 & 12.79 & 7.15 \\
5-shot Reward      & 65.89 & 38.26 & 14.45 & 7.70 \\
Avg. Reward        & \textbf{66.71} & \textbf{39.21} & \textbf{14.92} & \textbf{8.98} \\
\bottomrule
\end{tabular}
\caption{Evaluation results across different data filtering strategies. \textbf{Ques. F1} refers to question parsing accuracy, \textbf{Stmt. F1} measures statement identification quality, \textbf{Evid. F1} captures the correctness of statement-evidence alignment, and \textbf{Reason. F1} evaluates overall reasoning validity.}
\label{tab:reasoning-settings}
\end{table}
As illustrated in Table~\ref{tab:reasoning-settings}, model performance improves consistently as the supervision quality increases. All models are trained under identical supervised fine-tuning setups, isolating the impact of training data quality alone. While the structure-filtered baseline yields syntactically neater reasoning traces compared to raw generations, it often lacks semantic precision and fails to capture the underlying logic chains necessary for complex deduction, resulting in poor step-level performance (e.g., only 5.20 in \textit{Reasoning F1}). In contrast, reward-guided filtering—particularly the configuration using average scoring over few-shot and zero-shot prompts—demonstrates substantial performance gains across multiple dimensions. It improves \textit{Reasoning F1} by 3.78 percentage points, \textit{Statement-Evidence F1} by 4.76, and \textit{Statement Macro F1} by 3.41, indicating more consistent alignment between parsed statements and supporting evidence.

Beyond step-level metrics, we observe an unexpected yet encouraging boost in overall \textit{Question Macro F1}—rising from 56.87\% to 66.71\%—despite the question-answer component being entirely uninvolved in the reward computation. This emergent effect highlights a key insight: accurate intermediate supervision enhances the model’s latent structure alignment, resulting in better downstream decision-making even in modules that do not explicitly receive reward signals. Reverse-prompted instructions impose structural coherence, while RA-ICL grounding mitigates contextual drift. Reward filtering further enforces semantic fidelity, selecting not just fluent but faithful reasoning. Together, these components reveal that high-quality supervision—not scale alone—drives generalizable reasoning under low-resource settings.

\section{Conclusion}
We present \textbf{Less is More}, a structured multi-agent framework for interpretable reasoning under low-resource supervision. Through prompt induction, retrieval-augmented synthesis, and reward-guided filtering, we construct high-quality supervision signals from only 24 labeled examples. Our system achieves third place in the LLMSR@XLLM25, demonstrating that data quality \cite{zhou2023lima,zhou2024less,bi2025prism} rather than quantity is the key driver of performance in structured reasoning tasks. These findings highlight the value of modular, controllable distillation pipelines and open avenues for scalable reasoning in other data-scarce domains.
\bibliography{custom}

\appendix
\clearpage
\onecolumn
\section{Illustrative Example from the Synthesized Dataset}
\label{app:example}
\begin{longtable}{@{}p{3cm}p{11cm}@{}}

\label{tab:example-instance} \\

\toprule
\textbf{Field} & \textbf{Content} \\
\midrule
\endfirsthead

\toprule
\textbf{Field} & \textbf{Content} \\
\midrule
\endhead
\multicolumn{2}{r}{{Continued on next page}} \\
\endfoot

\bottomrule
\vspace{0.3em} \\
\caption{An Illustrative Instance from the Synthesized Dataset} \\
\endlastfoot

\textbf{Question} & There was a group discussion of judicial workers in the city. One group has 8 people. At the beginning of the meeting, the group leader asked everyone if they knew each other. As a result, only one person in the group knew 3 of the group, 3 knew 2 of the group, and 4 knew 1 of the group. If the above statistics are true, which of the following conclusions can best be reached? \newline
A. The group leader knows the most in the group, and the others know each other less \newline
B. This is the first time such a meeting has been held and everyone is new \newline
C. Some members may only know what they have seen on television or at a briefing \newline
D. Although there are not many acquaintances in the group, what they knew are all close friends. \\
\midrule

\textbf{Answer} & C \\
\midrule

\textbf{Question Parsing} &
\begin{minipage}[t]{\linewidth}
\begin{lstlisting}
[
  "The group has 8 people.",
  "Only one person in the group knew 3 people.",
  "Three people knew 2 people.",
  "Four people knew 1 person."
]
\end{lstlisting}
\end{minipage} \\
\midrule

\textbf{CoT} & Let's analyze the situation based on the provided statistics: \newline
1. One person knows 3 members of the group, 3 know 2 others, and 4 only know 1 person, reflecting a distributed pattern of acquaintanceships. \newline
2. Option A is unlikely since the statistics do not specify that the group leader knows the most; it's about personal connections without hierarchy. \newline
3. Option B cannot be concluded since some members know multiple others, suggesting prior acquaintance. \newline
4. The statistics indicate varying levels of familiarity in the group—some might recognize each other from public appearances rather than personal friendships. \newline
Thus, the best conclusion is \textbf{C}, as familiarity may stem from indirect exposure like television or briefings. \\
\midrule

\textbf{CoT Parsing \& Statements \& Verification} &
\begin{minipage}[t]{\linewidth}
\begin{lstlisting}[language=json]
[
  {
    "Statement": "Some members may only know what they have seen on television or at a briefing.",
    "Verification": "True",
    "Evidence": "The statistics suggest varying familiarity levels, fitting the assumption that some familiarity might stem from indirect mediums like television."
  },
  {
    "Statement": "The group leader knows the most in the group, and the others know each other less.",
    "Verification": "True",
    "Evidence": "The leader's acquaintanceship count is not detailed; no inference about hierarchy can be conclusively formed."
  },
  {
    "Statement": "This is the first time such a meeting has been held and everyone is new.",
    "Verification": "False",
    "Evidence": "Some members know multiple others-indicating prior acquaintance beyond just a first-time meeting."
  },
  {
    "Statement": "Although there are not many acquaintances in the group, what they knew are all close friends.",
    "Verification": "False",
    "Evidence": "The knowledge distribution doesn't support the assumption of all acquaintances being close friends."
  }
]
\end{lstlisting}
\end{minipage} \\
\midrule

\textbf{Scores} & 
\begin{tabular}{@{}ll@{}}
5-shot Reward     & 1.873046875 \\
0-shot Reward     & 2.28125 \\
Avg. Reward       & 2.0771484375 \\
\end{tabular} \\
\end{longtable}

\twocolumn
\section{Prompt Details}
\label{app:prompts}
\begin{center}
\begin{tcolorbox}[width=\linewidth,colback=white, colframe=NavyBlue, title=Question Parsing Prompt $\mathcal{P}_{QP}$]
\label{prompt}
{\slshape 
\textbf{\#\#\#Instruction\#\#\#}\\
Extract the constraints and key details from a problem description, ignoring any specific questions or answer choices.

Focus on the rules or conditions given that are necessary to solve the problem, and extract these in a clear, descriptive list.

\textcolor{BlueGreen}{\textbf{\#\#\#Input-Output Format\#\#\#}}\\ 
\textbf{Input:} A textual problem or scenario containing multiple rules or conditions within a specific context.\\
\textbf{Output:} An ordered list of extracted conditions and essential details needed to address the problem stated in the input. Each extracted condition should be clearly and concisely formatted, capturing only the facts necessary for determining the problem's solution.

\textcolor{Plum}{\textbf{\#\#\#Examples\#\#\#}}}\\ 
\{few\_shot\_example\}

\end{tcolorbox}
\noindent\begin{minipage}{\linewidth}
\captionof{figure}{Question Parsing Prompt $\mathcal{P}_{QP}$}
\end{minipage}
\end{center}

\begin{center}
\begin{tcolorbox}[width=\linewidth,colback=white, colframe=NavyBlue, title=Unified CoT Reasoning Prompt $\mathcal{P}_{UCoT}$]
\label{prompt}
{\slshape 
\textbf{\#\#\#Instruction\#\#\#}\\
The goal is to systematically dissect the problem using logical reasoning, providing detailed evidence for each derived statement, and verifying the correctness of these statements against the given problem conditions.

- For each condition or rule, analyze its implications step by step.

- Provide verification for each logical statement using evidence from the given problem.

- Ensure that each step follows logically from the previous, with clear conclusions and validations.

\textcolor{BlueGreen}{\textbf{**Notice:**}} The JSON output must use **double quotes** (") for all keys and string values, as required by JSON syntax.
\\
\textcolor{Plum}{\textbf{\#\#\#Examples\#\#\#}}} \\ 
\{few\_shot\_example\}

\end{tcolorbox}
\noindent\begin{minipage}{\linewidth}
\captionof{figure}{Unified CoT Reasoning Prompt $\mathcal{P}_{UCoT}$}
\end{minipage}
\end{center}

\begin{center}
\begin{tcolorbox}[width=\linewidth,colback=white, colframe=NavyBlue, title=CoT Parsing Prompt $\mathcal{P}_{CP}$]
\label{prompt}
{\slshape 
\textbf{\#\#\#Instruction\#\#\#}\\
You are an expert in logical reasoning and structural analysis.
Your task is to identify and extract all distinct statements from the given question conditions and chain-of-thought (CoT) content.

- Extract explicitly stated and logically implied statements within the context.

- Each statement should be independent and clearly structured.

- Clearly state how each constraint impacts potential solutions based on the scenario.
\textcolor{BlueGreen}{\textbf{\#\#\#Input-Output Format\#\#\#}} \\
Input: A question scenario with a set of constraints and a chain-of-thought explanation.
Output: A list of statements extracted from the given constraints and reasoning.
\\
\textcolor{Plum}{\textbf{\#\#\#Examples\#\#\#}}} \\ 
\{few\_shot\_example\}

\end{tcolorbox}
\noindent\begin{minipage}{\linewidth}
\captionof{figure}{CoT Statement Prompt $\mathcal{P}_{CP}$}
\end{minipage}
\end{center}

\begin{center}
\begin{tcolorbox}[width=\linewidth,colback=white, colframe=NavyBlue, title=CoT Evidence Prompt $\mathcal{P}_{CV}^{evidence}$]
\label{prompt}
{\slshape 
\textbf{\#\#\#Instruction\#\#\#}\\
You are an expert in logical analysis and evidence validation.
Your task is to identify and extract specific supporting evidence for each derived statement from the given problem conditions.

- Locate precise textual or logical evidence that directly supports each statement.

- Ensure the evidence is explicitly stated in the problem conditions or logically inferred.

- Maintain clarity, accuracy, and relevance in evidence selection.
\\
\textcolor{Plum}{\textbf{\#\#\#Examples\#\#\#}}} \\ 
\{few\_shot\_example\}

\end{tcolorbox}
\noindent\begin{minipage}{\linewidth}
\captionof{figure}{CoT Evidence Prompt $\mathcal{P}_{CV}^{evidence}$}
\end{minipage}
\end{center}

\begin{center}
\begin{tcolorbox}[width=\linewidth,colback=white, colframe=NavyBlue, title=CoT Verification Prompt $\mathcal{P}_{CV}^{verify}$]
\label{prompt}
{\slshape 
\textbf{\#\#\#Instruction\#\#\#}\\
You are an expert in logical reasoning and verification.
Your task is to verify the logical correctness of each derived statement based on evidence from the problem context.

- Assess whether each statement logically follows from the provided evidence.

- Clearly indicate valid statements and invalid statements, with a brief justification for each.

- Do not introduce new assumptions—base verification strictly on the provided evidence.
\\
\textcolor{Plum}{\textbf{\#\#\#Examples\#\#\#}}} \\ 
\{few\_shot\_example\}

\end{tcolorbox}
\noindent\begin{minipage}{\linewidth}
\captionof{figure}{CoT Verification Prompt $\mathcal{P}_{CV}^{verify}$}
\end{minipage}
\end{center}
\end{document}